\documentclass[11pt,a4paper]{article}
\usepackage[hyperref]{emnlp-ijcnlp-2019}
\usepackage{times}
\usepackage{latexsym}

\usepackage{url}

\aclfinalcopy 

\usepackage{xspace}
\usepackage[fleqn]{amsmath}
\usepackage{amssymb}

\usepackage{graphicx}

\usepackage{array}
\usepackage{booktabs}
\usepackage{morefloats}
\usepackage{float}
\usepackage{multirow}
\usepackage{subcaption}
\usepackage{xcolor}
\usepackage{pgfplots}
\usepackage{tikz}
\usetikzlibrary{patterns}
\usetikzlibrary{automata}
\usetikzlibrary{positioning}
\usepgfplotslibrary{groupplots}
\usepackage{helvet}
\usepackage[eulergreek]{sansmath}
\pgfplotsset{
  compat=1.14,
  tick label style = {font=\sffamily},
  every axis label = {font=\sffamily},
  legend style = {font=\sffamily},
  label style = {font=\sffamily}
}

\usepackage{xcolor}

\newcommand{\citepos}[1]{\citeauthor{#1}'s (\citeyear{#1})}

\usepackage[inline]{enumitem}

\newcommand{\colu}{}
\newcommand{\myparagraph}[1]{\smallskip\noindent{\textbf{#1.}~}}

\title{Leveraging Structural and Semantic Correspondence for Attribute-Oriented Aspect Sentiment Discovery}

\author{Zhe Zhang \\
  Watson Group\\
  IBM Corporation\\
  Research Triangle Park, NC 27703-9141 \\
  {\tt zhangzhe@us.ibm.com} \\\And
  Munindar P.~Singh \\
  Department~of Computer~Science \\
  North~Carolina~State~University \\
  Raleigh, NC 27695-8206 \\
  {\tt singh@ncsu.edu} \\}
  
\date{}

\begin{document}
\maketitle
\begin{abstract}

Opinionated text often involves attributes such as authorship and
location that influence the sentiments expressed for different
aspects. We posit that structural and semantic correspondence is both
prevalent in opinionated text, especially when associated with
attributes, and crucial in accurately revealing its latent aspect and
sentiment structure. However, it is not recognized by existing approaches.

We propose Trait, an unsupervised probabilistic model that discovers
aspects and sentiments from text and associates them with different
attributes. To this end, Trait infers and leverages structural and
semantic correspondence using a Markov Random Field. We show
empirically that by incorporating attributes explicitly Trait
significantly outperforms state-of-the-art baselines both by
generating attribute profiles that accord with our intuitions, as
shown via visualization, and yielding topics of greater semantic
cohesion.

\end{abstract}

\section{Introduction}
\label{sec:introduction}

Opinionated text is often associated with different
\emph{attributes}---latent variables that serve as reference frames
relative to which the underlying aspects and sentiments are expressed.
Common attributes in consumer reviews include author type (e.g.,
business traveler or tourist for hotel reviews; location for reviews
of music \cite{McDermott:16}; culture on reviews of food
\cite{Bahauddin:15}). Whereas current approaches consider attributes
in a one-off manner in each application, we posit that attributes can
be systematically extracted if we can properly capture the structural
and semantic correspondence that is prevalent in opinionated text. We
claim that ignoring attributes may lead to biased inference on aspects
and sentiments. As evidence, we demonstrate an approach that
outperforms the state of the art and yields intuitive and cohesive
topics.

We propose Trait, a general model for discovering attribute-oriented
aspects and sentiments from text. By incorporating attributes, Trait
automatically generates profiles that describe attributes in terms of
sentiments and aspects. To leverage structural and semantic
correspondence, Trait applies a Markov Random Field as regularization
over sentences during inference. We evaluate Trait on four datasets
from two domains and consider three attributes. Trait successfully
discovers aspects associated with sentiments; the generated word
clusters are more cohesive than the state-of-the-art baselines; the
generated attribute profiles are well correlated with ground truth.

\begin{figure*}[ht!]
\centering
\includegraphics[width=1
\textwidth]{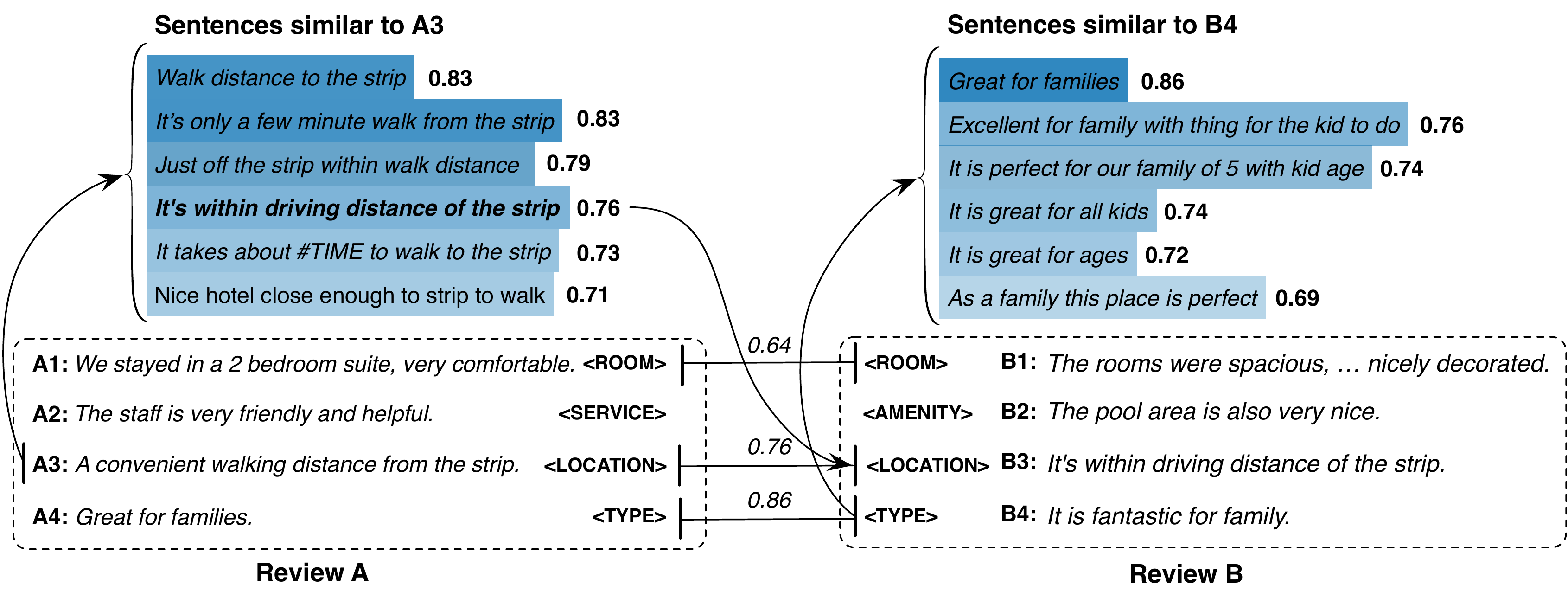}
\caption{A motivating example.}
\label{fig:example}
\end{figure*}

\myparagraph{Motivating Example} Figure~\ref{fig:example} presents two
hotel reviews from TripAdvisor. We manually assign aspect labels for
sentences and calculate pairwise cosine similarity between sentences
using sentence embedding from a pretrained sentence encoding model
\cite{USE:2018}.

Review~A and Review~B, which mention aspects \emph{Room},
\emph{Location}, and \emph{Type}, exhibit structural and semantic
correspondence. We posit that the correspondence of \emph{Location}
and \emph{Type} is a result of the attribute value, Las Vegas, common
to the two reviews. \emph{Location} is a crucial aspect for hotels in
Las Vegas. On a randomly selected set of 5,000 hotel reviews for Las
Vegas, we observe that 4,281 sentences from 2,624 reviews mention the
location ``Strip.'' Using a similarity threshold of 0.6, we obtain
1,929 sentences similar to sentence A3 from 1,519 reviews, including
sentence B3 in Review~B. We obtain 133 sentences similar to sentence
B4 from 128 reviews including A4 in Review~A. Figure~\ref{fig:example}
shows some of these sentences. Likewise, using authorship as an
attribute, we observe that users stick to their writing styles. For
example, in hotel reviews, some users describe the condition of a room
and others share travel tips.

\myparagraph{Contributions and Novelty} Our contributions include: (1)
a general model that generates attribute profiles associating aspects
and sentiments with attributes in text; (2) empirical results
demonstrating the benefit of incorporating attributes on a model's
quality; and (3) empirical results demonstrating generalizability
using diverse attributes and the quality of the generated attribute
profiles.

Trait's novelty lies in its ability to accommodate attributes. First,
it is general across attributes as opposed to being limited to
predefined attributes. Second, the handling of attributes means that
Trait avoids overfitting to the more prevalent attributes in a
dataset. That is, Trait can learn a more refined conditional
probability distribution that incorporates specific attributes than
otherwise possible. Ignoring the observable attribute variables would
relax the constraints on the distribution, meaning that the learned
approximate distribution would be biased toward the majority attribute.

\myparagraph{Summary of Findings} We demonstrate that incorporating
attributes into generative models provides a superior, more refined
representation of opinionated text. The resulting model generates
topics with high semantic cohesion. We show that Markov Random Field
can be used for effectively capturing structural and semantic
correspondence.

\section{Related Work}

Generative probabilistic modeling has been widely applied for
unsupervised text analysis. Given the observed variables, e.g., tokens
in documents, a generative probabilistic model defines a set of
dependencies between hidden and observed variables that encodes
statistical assumptions underlying the data. Latent Dirichlet Allocation
(LDA) \cite{Blei:2003}, a well-known topic model, represents a
document as a mixture of topics, each topic being a multinomial
distribution over words. The learning process approximates the topic
and word distributions based on their co-occurrence in documents.

Many efforts guide the topics learned by incorporating additional
information. \citepos{Rosen-Zvi:2004} Author Topic model (AT) captures
authorship by building a topic distribution for each author. When
generating a word in a document, AT conditions the probability of
topic assignment on the author of the document. \citepos{KimSHH:2012}
model captures entities mentioned in documents and models the
probability of generating a word as conditioned on both entity and
topic. \citet{DiaoJ:2013} jointly model topics, events, and users on
Twitter. Trait goes beyond these models by incorporating sentiments
and attributes in a flexible way, which eliminates the model's
dependency on specific attribute types.

Several probabilistic models tackle opinionated text.
\citet{TitovWWW:08} handle global and local topics in documents. JST
\cite{JST:2012} and ASUM \cite{Jo:2011} model a review via multinomial
distributions of topics and sentiments used to condition the
probability of generating words. \citet{Kim:2013} extend ASUM's
probabilistic model to discover a hierarchical structure of
aspect-based sentiments. \citepos{WangC0:16} topic model discovers
aspect, sentiment, and both general and aspect-specific opinion words.
Whereas these models identify aspects and sentiments, they disregard
attribute information.

\citet{Titov:08} discover topics using aspect ratings provided by
reviewers. \citepos{Mukherjee:14} JAST considers authors during aspect
and sentiment discovery. \citepos{PoddarHL:17} AATS jointly considers
author, aspect, sentiment, and the nonrepetitive generation of aspect
sequences via a Bernoulli process. \citet{Zhang&Singh:18}'s model
jointly captures aspect, sentiment, author, and discourse relations.
Trait is novel in that, unlike the above models, it is not tied to a
specific attribute.

\section{Model and Inference}

We now introduce Trait's model and inference mechanism.

\subsection{Sentence Embeddings}

Measuring semantic similarity between sentences is integral to
capturing the structural and semantic correspondence among reviews:
high similarity indicates a high degree of correspondence.
\citet{USE:2018} propose a pretrained sentence encoding model,
Universal Sentence Encoder (USE). USE is based on \citepos{Vaswani:17}
attention-based neural network. \citet{Perone:18} show USE yields the
best results among sentence embedding techniques on semantic
relatedness and textual similarity tasks. Trait adopts USE to generate
sentence embeddings and cosine similarity to measure semantic
similarity between sentences.

\subsection{Structural and Semantic Correspondence}
A Markov Random Field (MRF) defines a joint probability distribution
over a set of variables given the dependencies based on an undirected
graph. The joint distribution is a factorized product of potential
functions.

To capture structural and semantic correspondence, Trait defines an
MRF over latent aspects of sentences. Given a set of reviews
$D_a$ sharing a common attribute $a$, for sentence $l$ in $D_a$, Trait
creates its corresponding sentence set $L$ by adding sentence $l_i$ in
$D_a$ if the semantic similarity between sentence $l_i$ and $l$ is
larger than a preset threshold $\rho$. For each pair of $l$ and $l_i$,
Trait creates an undirected edge between the aspects associated with
the two sentences ($t_{l}$, $t_{li}$). To promote $l$ and sentences in
$L$ having a high probability of associating with the same aspect,
Trait defines a binary edge potential,
$\exp\{\mathcal{I}(t_{l},t_{li})\}$, where $\mathcal{I}(\cdot)$ is an
indicator function. This binary potential produces a large value if
the two sentences have the same aspect and a small value otherwise. Given attribute $a$, sentiment $s$, and a document consisting of $N$ sentences, Trait computes the joint probability of aspect assignments of sentences as:

{\setlength{\mathindent}{0cm}
\begin{align}
\label{eq:correspondence}
\begin{split}
p(\boldsymbol{t}|\psi_{s,a}, \lambda)=&\prod_{l}^{N}p(t_l|\psi_{s,a})\\&\exp\bigg\{\lambda\frac{\sum\nolimits_{(t_{l}, t_{l_i})\in \mathcal{E}_l} \mathcal{I}(t_{l}=t_{l_i})}{\lvert\mathcal{E}_l\rvert}\bigg\},
\end{split}
\end{align}}

{\noindent}where $\psi_{s,a}$ is the aspect distribution given sentiment $s$ and attribute $a$; parameter $\lambda$
controls the reinforcing effects of correspondence
regularization; and $\mathcal{E}_l$ is the set of undirected edges for
$l$.


\subsection{Generative Process}

To capture the desired associations, given an attribute type, Trait
generates a mixture over sentiments and aspects for each attribute
value. Trait assumes that reviews are mixtures of sentiments and
considers sentences the basic unit for a sentiment-aspect pair.

\begin{figure}[htb]
\centering
\includegraphics[width=0.48
\textwidth]{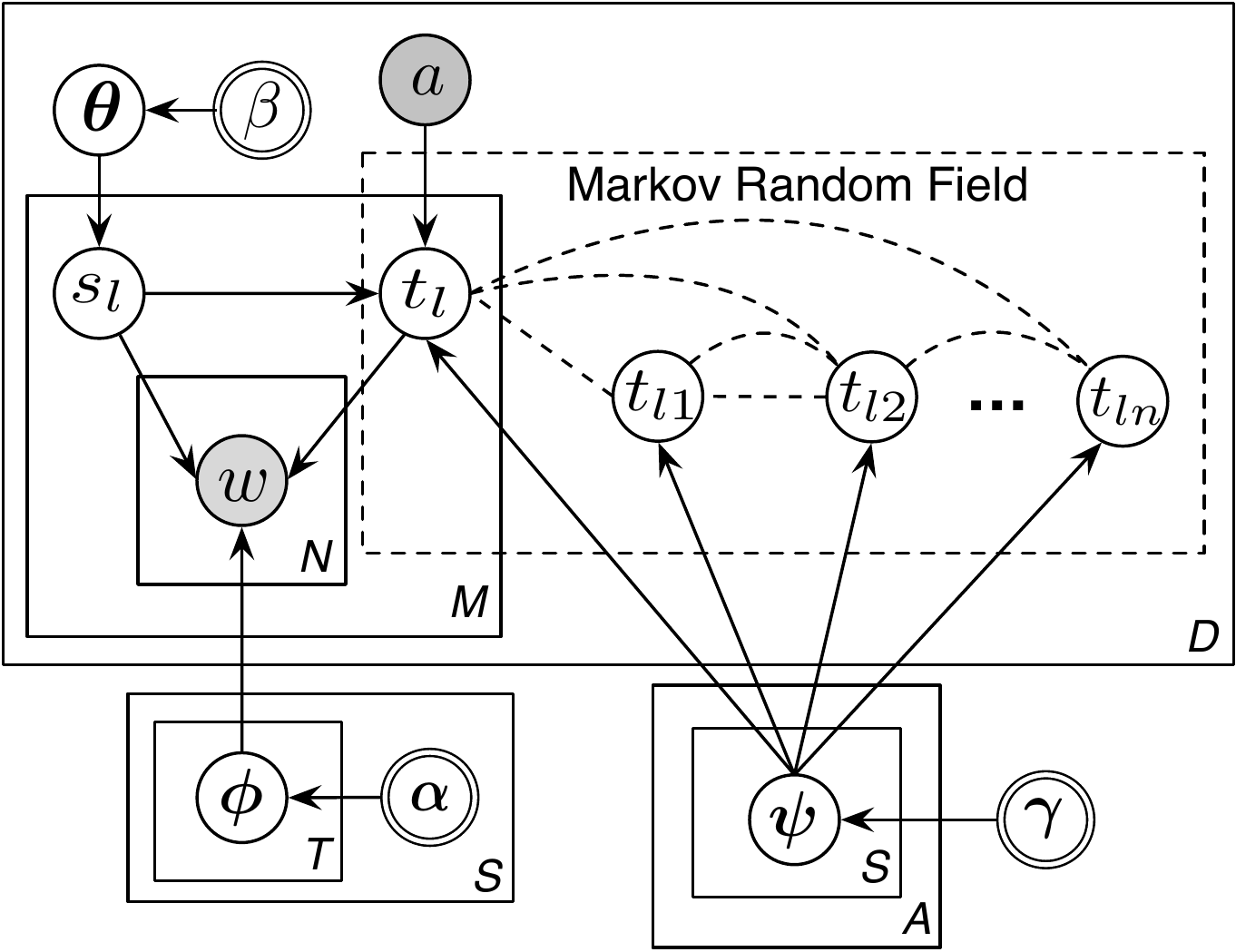}
\caption{Generative process of Trait.}
\label{fig:graphical_model}
\end{figure}
\setlength{\textfloatsep}{0pt}
\setlength{\abovedisplayskip}{-2pt}

Figure~\ref{fig:graphical_model} shows Trait's model. Hyperparameter
$\alpha$ is the Dirichlet ($\mathit{Dir}(\cdot)$) prior of the word
distribution $\phi$; $\beta$ is the Dirichlet prior of the sentiment
distribution $\theta$; and $\gamma$ is the Dirichlet prior of the
aspect distribution $\psi$. Given a set of reviews $D$ associated with
a set of attribute values $A$ over a set of aspects $T$ and a set of
sentiments $S$, each review contains $M$ sentences and each sentence
contains $N$ words. Trait's generative process is as follows.

First, for each pair of aspect $t$ and sentiment $s$, draw a word
distribution $\phi_{t,s} \sim \mathit{Dir}(\alpha)$. Second, for each
attribute value $a$ and each sentiment $s$, draw an aspect
distribution $\psi_{s,a} \sim \mathit{Dir}(\gamma)$. Third, given a
review $d$ with attribute $a$, draw a sentiment distribution $\theta_d
\sim \mathit{Dir}(\beta)$, and for each sentence in $d$, (1) choose a
sentiment $s \sim \mathit{Multinomial}(\theta_d)$; (2) given $s$,
choose an aspect $t \sim \mathit{Multinomial}(\psi_{s,a})$; (3) given
$t$ and $s$, sample word $w \sim \mathit{Multinomial}(\phi_{t,s})$.

Trait estimates $p(\boldsymbol{s},\boldsymbol{t}|\boldsymbol{w},a)$,
the posterior distribution of latent variables, sentiments
$\boldsymbol{s}$, and aspects $\boldsymbol{t}$, given all words used
in reviews involving attribute $a$. We factor the joint probability of
assignments of sentiments, aspects, and words for $a$:

{
\begin{align}
\label{eq:joint}
p(\boldsymbol{s},\boldsymbol{t},\boldsymbol{w}|a)  =
p(\boldsymbol{w}|\boldsymbol{s},\boldsymbol{t}) p(\boldsymbol{t}|\boldsymbol{s},a) p(\boldsymbol{s}).
\end{align}}

\subsection{Inference}

We use collapsed Gibbs sampling \cite{Liu:1994} for posterior
inference. By integrating over $\Phi =
\{\boldsymbol{\phi_i}\}_{i=1}^{S\times T}$, we obtain
Equation~\ref{eq:joint}'s first term (Section~\ref{subsec:parameter}
explains $\alpha_v$).

{
\begin{flalign}
\label{eq:wsz}
\medmuskip=0mu
\thinmuskip=0mu
\thickmuskip=0mu
\nulldelimiterspace=-1pt
\scriptspace=-1pt
\hspace*{-2em}
\begin{split}
&p(\boldsymbol{w}|\boldsymbol{s},\boldsymbol{t},\boldsymbol{\alpha})=\int \!  p(\boldsymbol{w}|\boldsymbol{s},\boldsymbol{t},{\Phi}) p({\Phi}|\boldsymbol{\alpha}) \, \mathrm{d}{\Phi} \\
&=\left(\frac{\Gamma(\sum_{v=1}^{W}\alpha_{v})}{\prod_{v=1}^{W}\Gamma(\alpha_{v})}\right)^{S \times T}\!\times\prod_{s=1}^{S} \prod_{t=1}^{T} \frac{\prod_{v=1}^{W} \Gamma(n_{s,t}^{v}+\alpha_{v})}{\Gamma\big[\sum_{v=1}^{W}(n_{s,t}^{v} +\alpha_{v} )\big]},
\end{split}
\end{flalign}}

{\noindent}where $W$ is the vocabulary size; $n_{s,t}^{v}$ is the
number of occurrences of word $v$ assigned to sentiment $s$ and aspect
$t$; and $\Gamma(\cdot)$ is the Gamma function.

Next, by integrating over $\Psi_{a} =
\{\boldsymbol{\psi_i}\}_{i=1}^{S}$, we calculate the second term in
Equation~\ref{eq:joint} as (Section~\ref{subsec:parameter} explains
$\gamma_t$):

\begin{align}
\label{eq:szu}
\medmuskip=0mu
\thinmuskip=0mu
\thickmuskip=0mu
\hspace*{-2em}
\begin{split}
&p(\boldsymbol{t}|\boldsymbol{s},\boldsymbol{\gamma},a) =\int \!
p(\boldsymbol{t}|\boldsymbol{s},{\Psi_{a}}, a) p({\Psi_{a}}|\boldsymbol{\gamma}) \, \mathrm{d}{\Psi_{a}} \\
&=\left(\frac{\Gamma(\sum_{t=1}^{T}\gamma_{t})}{\prod_{t=1}^{T}\Gamma(\gamma_{t})}\right)^{S}\times\prod_{s=1}^{S} \frac{\prod_{t=1}^{T} \Gamma(n_{s,a}^{t}+\gamma_{t})}{\Gamma \big[\sum_{t=1}^{T}(n_{s,a}^{t} +\gamma_{t} )\big]}\\&
\times \prod_{m=1}^{M} exp\bigg\{\lambda\frac{\sum\nolimits_{t_{j}\in L_m} \mathcal{I}(t_{j}=t)}{\lvert L_m\rvert}\bigg\},
\end{split}
\end{align}

{\noindent}where $n_{s,a}^{t}$ equals the number of sentences in
reviews associated with attribute $a$, sentiment $s$, and aspect $t$;
$M$ is the number of sentences in reviews; $L_m$ is the set of sentences corresponding to sentence $m$.

Similarly, for the third term in Equation~\ref{eq:joint}, by
integrating over $\Theta = \{\boldsymbol{\theta_i}\}_{i=1}^{D}$, we
obtain (Section~\ref{subsec:parameter} explains $\beta_s$):

{
\begin{align}
\label{eq:z}
\medmuskip=0mu
\thinmuskip=0mu
\thickmuskip=0mu
\begin{split}
&p(\boldsymbol{s}|\boldsymbol{\beta}) =\int \!  p(\boldsymbol{s}|{\Theta}) p(\Theta|\boldsymbol{\beta}) \, \mathrm{d}{\Theta} \\
&=
\left(\frac{\Gamma(\sum_{s=1}^{S}\beta_{s})}{\prod_{s=1}^{S}\Gamma(\beta_{s})}\right)^{D}\times\prod_{d=1}^{D} \frac{\prod_{s=1}^{S} \Gamma(n_{d}^{s}+\beta_{s})}{\Gamma\big[\sum_{s=1}^{S}(n_{d}^{s} +\beta_{s} )\big]},
\end{split}
\end{align}}

{\noindent}where $D$ is the number of reviews; $n_{d}^{s}$ is the
number of times that a sentence from review $d$ is associated with
sentiment $s$; and $n_{d}$ is the number of sentences in review $d$.

For each sweep of a Gibbs iteration, we sample latent aspect $t$ and
sentiment $s$ as follows:

{\setlength{\mathindent}{0cm}
\begin{align}
\label{eq:Gibbs}
\begin{split}
&p(s_i=s, t_i=t|\boldsymbol{s_{-i}},\boldsymbol{t_{-i}},\boldsymbol{w},a)\\& \propto \frac{n_{s,a,-i}^{t}+\gamma_{t}}{\sum_{t=1}^{T} (n_{s,a,-i}^{t} +\gamma_{t})}
\!\times\! \frac{n_{d,-i}^{s}+\beta_{s}}{\sum_{s=1}^{S}(n_{d,-i}^{s} +\beta_{s})} \\& \times \frac{\prod_{v \in W_i}\prod_{c=0}^{C_{v}^{i}-1} (n_{s,t,-i}^{v} + \alpha_{v} + c)}{\prod_{c=0}^{C_i-1}(n_{s,t,-i} + \sum_{v=1}^{W} \alpha_{v} +c)}\\&
\times exp\bigg\{\lambda\frac{\sum\nolimits_{t_{j}\in L_i} \mathcal{I}(t_{j}=t)}{\lvert L_i\rvert}\bigg\},
\end{split}
\end{align}
}

where $n_{s,a}^{t}$ is the number of sentences from reviews associated
with attribute $a$, sentiment $s$, and aspect $t$; $n_{d}^{s}$ is the
number of sentences from review $d$ associated with sentiment $s$; $W_i$ is the set of words in sentence $i$. $C_{v}^{i}$ is the count of word $v$ in sentence $i$; $C_i$ is the
number of words in sentence $i$; $n_{s,t}^{v}$ is the number of words
$v$ assigned sentiment $s$ and aspect $t$; $n_{s,t}$ is the number of
words assigned sentiment $s$ and aspect $t$ in all reviews; $L_i$ is
the set of corresponding sentences of sentence $i$; and an index of
$-i$ indicates excluding sentence $i$ from the count.

Equations~\ref{eq:varphi}, \ref{eq:psi}, and \ref{eq:pi},
respectively, approximate the probabilities of word $w$ occurring
given sentiment $s$ and aspect $t$; of aspect $t$ of a sentence
occurring given sentiment $s$ and attribute $a$; of sentiment $s$
occurring given document $d$.

{

\begin{align}
\label{eq:varphi}
\begin{split}
\phi_{s,t,w} = \frac{n_{s,t}^{w} + \alpha_{w}}{\sum_{v=1}^{W} (n_{s,t}^{w} +  \alpha_{v})},
\end{split}
\end{align}

\begin{align}
\label{eq:psi}
\begin{split}
\psi_{s,t,a} = \frac{n_{s,a}^{t} + \gamma_{t}}{\sum_{t=1}^{T} (n_{s,a}^{t} + \gamma_{t})},
\end{split}
\end{align}

\begin{align}
\label{eq:pi}
\begin{split}
\theta_{d,s} = \frac{n_{d}^{s} + \beta_{s}}{ \sum_{s=1}^{S} (n_{d}^{s} +\beta_{s})}.
\end{split}
\end{align}

}

The generalized P\'{o}lya Urn model \cite{Mahmoud:08} has been used for
encoding word co-occurrence information into topic models. Consider an
urn containing a mixture of balls, each of which is tagged with a
term, for each sampling sweep we draw a ball from the urn. In a
standard P\'{o}lya Urn model, as used in LDA, we return the ball to the
urn with another ball tagged with the same term. This process provides
burstiness of the probability of seeing a term but ignores the
covariance. The probability increase of one term decreases the
probability of the other words. In the generalized P\'{o}lya Urn model,
when a ball is drawn from the urn, we replace it with two new balls
with a set of balls tagged with related terms. Similar to previous
models \cite{MimnoWTLM:11, ChenMLHCG:13, Fei:14, Zhang&Singh:18}, to
increase the probability of having semantically related words appear in
the same topic, Trait applies a generalized P\'{o}lya Urn model in
each Gibbs sweep and uses weight $\varepsilon$ to promote related
words based on the cosine similarity between their Word2Vec
\cite{MikolovSCCD:13} embeddings.

\section{Evaluation}
\label{sec:evaluation}

To assess Trait's effectiveness, we select the hotel and restaurant domains
and prepare four review datasets associated with three attributes:
author, trip type, and location. HotelUser, HotelType, and HotelLoc
are sets of hotel reviews collected from TripAdvisor. HotelUser
contains 28,165 reviews posted by 202 randomly selected reviewers,
each of whom contributes at least 100 hotel reviews. HotelType
contains reviews associated with five trip types including business,
couple, family, friend, and solo. HotelLoc contains a total of 136,446
reviews about seven US cities, split approximately equally. ResUser is
a set of restaurant reviews from Yelp Dataset Challenge
\shortcite{Yelp:2019}. It contains 23,874 restaurant reviews posted by
144 users, each of whom contributes at least 100 reviews.
Table~\ref{tab:data} summarizes our datasets. Datasets and source code
are available for research purposes \cite{TraitCode:2019}.

\begin{table}[htb]
\centering
\caption{Summary of the evaluation datasets.}
\label{tab:data}
\scalebox{0.75} {
\begin{tabular}{@{~~}l@{~~~}r@{~~~}r@{~~~}r@{~~~}r@{~~}}
 \toprule
 \colu{Statistic} & \colu{HotelUser} &  \colu{ResUser}&  \colu{HotelType}&  \colu{HotelLoc}\\ \midrule \\[-1.5ex]
\# of reviews & 28,165 & 23,873 &  22,984 & 136,446 \\ [0.5ex]
\# of sentences & 362,153 & 276,008 & 302,920 & 1,428,722\\ [0.5ex]
Sentence/Review & 13 & 12 & 13 & 10\\ [0.5ex]
Words/Sentence & 8 & 7 & 7 & 7\\ [0.5ex] \bottomrule
\end{tabular}
}
\end{table}

We remove stop words and HTML tags, expand typical abbreviations, and
mark special named entities using a rule-based algorithm (e.g.,
replace a URL by \#LINK\# and replace a monetary amount by \#MONEY\#)
and the Stanford named entity recognizer \cite{Finkel:05}. We use
\citepos{Porter:2006} stemming algorithm. To handle negation, for any
word pair whose first word is \emph{no}, \emph{not}, or
\emph{nothing}, we replace the word pair by a negated term, e.g.,
producing \emph{not\_work} and \emph{not\_quiet}. Finally, we split
each review into constituent sentences.

\subsection{Parameter Settings}
\label{subsec:parameter}

Trait includes three manually tuned hyperparameters that have a
smoothing effect on the corresponding multinomial distributions.
Hyperparameter $\alpha$ is the Dirichlet prior of the word
distribution. We use asymmetric priors based on a sentiment lexicon.
Table~\ref{tab:lexicon} shows Trait's sentiment word list as prior
knowledge to set asymmetric priors. This list extends
\citepos{TurneyL:03} list with additional general sentiment words. For
any word in the positive list, we set $\alpha$ to 0 if this word
appears in a sentence assigned a negative sentiment, and to 5 if this
word appears in a sentence assigned a positive sentiment, and
conversely for words in the negative list. For all remaining words, we
set $\alpha$ to 0.05. We set hyperparameter $\beta$, the Dirichlet
prior of the sentiment distribution, to 5 for both sentiments. We set
hyperparameter $\gamma$, the Dirichlet prior of the aspect
distribution, to 50/$T$ for all models, where $T$ is the number of
aspects. We set the reinforcement weight of structural and semantic
correspondence $\lambda$ to 1.0; sentence semantic similarity $\rho$
to 0.7; and, related word promoting weights to 0.3 for hotel reviews
and 0.1 for restaurant reviews.

\begin{table}
\centering
\caption{Lists of sentiment words.}
\label{tab:lexicon}
\scalebox{0.98} {
\begin{tabular}{p{0.95\columnwidth}}
 \toprule
 \textbf{Positive}\\\midrule
  amazing, attractive, awesome, best, comfortable, correct,
enjoy, excellent, fantastic, favorite, fortunate, free, fun,
glad, good, great, happy, impressive, love, nice, not\_bad, perfect,
positive, recommend, satisfied, superior, thank, worth  \\ [0.5ex]  \midrule
 
\textbf{Negative}\\ \midrule
annoying, bad, complain, disappointed, hate, inferior, junk, mess, nasty, negative, not\_good, not\_like, not\_recommend, not\_worth, poor, problem, regret, slow, small, sorry, terrible, trouble, unacceptable, unfortunate, upset, waste, worst, worthless, wrong\\ [0.5ex]
\bottomrule
\end{tabular}
}
\end{table}

\subsection{Quantitative Evaluation}
\label{subsec:qualitative}

Whether topics (word clusters) are semantically cohesive is crucial in
assessing topic modeling approaches. As in previous studies
\cite{NguyenBDJ:15,NguyenBLSR:15,YangBR:17}, we adopt Normalized
Pointwise Mutual Information (NPMI) \cite{LauNB:14} and W2V
\cite{OCallaghanGCC:15} as our evaluation metrics. Higher NPMI and W2V
scores indicate greater semantic cohesion. We compare Trait with four
baselines: AT, JST, ASUM, and AATS. We perform our evaluation on
HotelUser and ResUser based on the top 20 words in each
sentiment-aspect pair. We split data into five folds and use training
split to train all models. For each number of aspects, we conduct a
two-tailed paired t-test for each of the pairwise comparisons.
Throughout, $*$, $\dagger$, and $\ddagger$ indicate significance at
0.05, 0.01, and 0.001, respectively.

Table~\ref{tab:cohesion_trip} shows average NMPI and W2V scores for
different numbers of aspects. AT performs worst, possibly due to
missing conditions on sentiments. ASUM and JST are comparable. Trait
outperforms all others, with the highest NMPI and W2V scores for each
number of aspects. Table~\ref{tab:cohesion_yelp} shows similar
conclusions for restaurant reviews.

For both datasets, Trait's improvements of topic coherence over
baseline models are statistically significant for HotelUser (p $<$
0.001) and ResUser (p $=$ 0.002). Trait allows reviews written by the
same or similar authors to have idiosyncratic preferences over aspects
and sentiments. Trait assigns aspects to sentences by sampling
attribute-specific aspect distributions. These distributions are
regularized by the Markov Random Fields. Sentences with a high degree of
correspondence have a high probability to be assigned the same aspects.

\begin{table}[htb]
\centering
\caption{Topic coherence: Hotel reviews.}
\label{tab:cohesion_trip}
\scalebox{0.95} {\footnotesize
\begin{tabular}{lrrrrrr}\toprule
NPMI&T=10&T=20&T=30&T=40&T=50&T=60\\\midrule
AT&3.64&4.04&4.37&4.49&4.86&5.14\\
AATS&5.63&9.08&10.41&10.78&11.05&11.00\\
JST&8.99&10.78&11.45&11.54&11.56&11.46\\
ASUM&9.48&10.64&11.02&11.33&11.39&11.56\\
Trait&\textbf{15.50}\rlap{\textsuperscript{\textdaggerdbl}}&\textbf{16.91}\rlap{\textsuperscript{\textdaggerdbl}}&\textbf{17.31}\rlap{\textsuperscript{\textdaggerdbl}}&\textbf{17.32}\rlap{\textsuperscript{\textdaggerdbl}}&\textbf{16.46}\rlap{\textsuperscript{\textdaggerdbl}}&\textbf{15.34}\rlap{\textsuperscript{\textdaggerdbl}}\\\midrule

W2V&T=10&T=20&T=30&T=40&T=50&T=60\\\midrule
AT&0.10&0.10&0.10&0.09&0.09&0.09\\
AATS&0.13&0.16&0.18&0.18&0.18&0.18\\
JST&0.15&0.18&0.18&0.19&0.19&0.19\\
ASUM&0.17&0.18&0.18&0.18&0.18&0.18\\
Trait&\textbf{0.33}\rlap{\textsuperscript{\textdaggerdbl}}&\textbf{0.35}\rlap{\textsuperscript{\textdaggerdbl}}&\textbf{0.35}\rlap{\textsuperscript{\textdaggerdbl}}&\textbf{0.34}\rlap{\textsuperscript{\textdaggerdbl}}&\textbf{0.32}\rlap{\textsuperscript{\textdaggerdbl}}&\textbf{0.31}\rlap{\textsuperscript{\textdaggerdbl}}\\\bottomrule
\end{tabular}
}
\end{table}

\begin{table}[htb]
\centering
\caption{Topic coherence: Restaurant reviews.}
\label{tab:cohesion_yelp}
\scalebox{0.95} {\footnotesize
\begin{tabular}{lrrrrrr}\toprule
NPMI&T=10&T=20&T=30&T=40&T=50&T=60\\\midrule
AT&5.64&5.21&5.30&5.65&6.54&7.94\\
AATS&6.05&8.02&9.03&9.35&9.90&9.95\\
JST&9.46&11.13&11.73&11.92&12.14&\textbf{12.31}\\
ASUM&8.81&9.7&9.92&10.09&10.07&10.04\\
Trait&\textbf{11.27}\rlap{\textsuperscript{\textdaggerdbl}}&\textbf{13.02}\rlap{\textsuperscript{\textdaggerdbl}}&\textbf{13.62}\rlap{\textsuperscript{\textdaggerdbl}}&\textbf{13.18}\rlap{\textsuperscript{\textdaggerdbl}}&\textbf{12.36}\rlap{\textsuperscript{}}&11.95\rlap{\textsuperscript{}}\\\midrule

W2V&T=10&T=20&T=30&T=40&T=50&T=60\\\midrule
AT&0.16&0.15&0.14&0.13&0.13&0.14\\
AATS&0.11&0.14&0.16&0.17&0.18&0.18\\
JST&0.21&0.21&0.20&0.20&0.20&0.19\\
ASUM&0.20&0.19&0.18&0.18&0.17&0.17\\
Trait&\textbf{0.25}\rlap{\textsuperscript{\textdaggerdbl}}&\textbf{0.26}\rlap{\textsuperscript{\textdaggerdbl}}&\textbf{0.25}\rlap{\textsuperscript{\textdaggerdbl}}&\textbf{0.24}\rlap{\textsuperscript{\textdaggerdbl}}&\textbf{0.24}\rlap{\textsuperscript{\textdaggerdbl}}&\textbf{0.24}\rlap{\textsuperscript{\textdaggerdbl}}\\\bottomrule
\end{tabular}
}
\end{table}

\subsection{Sentiment Classification}
\label{subsec:sentiment}

Automatically detecting the sentiment of a document is an important
task in sentiment analysis. We compare Trait with JST, ASUM, and AATS
for document-level sentiment classification using HotelUser and
ResUser. We use integer ratings of reviews to collect ground-truth
labels. Reviews with ratings at three and above are labeled as
positive and the rest are labeled as negative. Note that our datasets
are imbalanced. We conduct five-fold cross-validation with the two-tailed
paired t-test. For each user, we use 80\% of reviews for training and
20\% for testing. For evaluation metrics, we adopt accuracy (Acc) and
area under the curve (AUC) of the Receiver Operating Characteristic (ROC) curve. ROC plots the true positive rate
against the false positive rate. AUC-ROC is a standard metric for
evaluating classifiers' performance on imbalanced data.

\begin{table}[htb]
\centering
\caption{Accuracy and AUC of sentiment classification on hotel reviews.}
\label{tab:compare_overall_trip}
\scalebox{1} {\footnotesize
\begin{tabular}{lrrrrrr}\toprule
\multicolumn{1}{c}{\multirow{2}{*}[-.3em]{}}&\multicolumn{2}{c}{T=20}&\multicolumn{2}{c}{T=40} &\multicolumn{2}{c}{T=60}\\
 \cmidrule(r){2-3} \cmidrule(r){4-5} \cmidrule(r){6-7} 
  &Acc&AUC&Acc&AUC&Acc&AUC\\\midrule
AATS&0.79&0.45&0.82&0.48&0.84&0.48\\
JST&0.61&0.82&0.64&0.83&0.67&0.84\\
ASUM&0.80&0.83&0.84&0.84&0.87&0.83\\
Trait&\textbf{0.85}\rlap{\textsuperscript{\textdagger}}&\textbf{0.86}\rlap{\textsuperscript{*}}&\textbf{0.87}\rlap{\textsuperscript{*}}&\textbf{0.85}\rlap{\textsuperscript{*}}&\textbf{0.88}&\textbf{0.86}\rlap{\textsuperscript{\textdagger}}\\\bottomrule
\end{tabular}
}
\end{table}

Table~\ref{tab:compare_overall_trip} reports the results of sentiment
classification on hotel reviews. AATS achieves better accuracy but
worse AUC scores than JST. ASUM yields better accuracy and comparable
AUC scores compared with JST. Trait consistently outperforms all
baseline models given different aspect numbers with an average gain in
accuracy of 3\%. Incorporating attributes and structural and semantic
correspondence into conditional probability distributions greatly
benefit the model in capturing dependencies among attributes,
aspects, and sentiments.

\begin{table}[htb]
\centering
\caption{Accuracy and AUC of sentiment classification on restaurant reviews.}
\label{tab:compare_overall_yelp}
\scalebox{1} {\footnotesize
\begin{tabular}{lrrrrrr}\toprule
\multicolumn{1}{c}{\multirow{2}{*}[-.3em]{}}&\multicolumn{2}{c}{T=20}&\multicolumn{2}{c}{T=40} &\multicolumn{2}{c}{T=60}\\
 \cmidrule(r){2-3} \cmidrule(r){4-5} \cmidrule(r){6-7} 
  &Acc&AUC&Acc&AUC&Acc&AUC\\\midrule
AATS&0.79&0.47&0.80&0.48&0.82&0.50\\
JST&0.59&0.71&0.61&0.73&0.64&0.73\\
ASUM&0.80&0.78&0.84&0.78&0.87&0.74\\
Trait&\textbf{0.86}\rlap{\textsuperscript{\textdagger}}&\textbf{0.79}&\textbf{0.87}\rlap{\textsuperscript{\textdagger}}&0.78&\textbf{0.88}&\textbf{0.79}\rlap{\textsuperscript{*}}\\\bottomrule
\end{tabular}
}

\end{table}

\subsection{Attribute Profile}
\label{subsec:author-profile}

Given reviews with selected attributes, we
expect Trait to generate profiles representing the characteristics
associated with those attributes. To evaluate profiles, we run Trait
on HotelUser, HotelLoc, and HotelType, associated with three
attributes: authors, locations, and trip types, respectively.

\subsubsection{Summarization}
\label{subsubsec:summarization}

Trait outputs profiles that summarize attributes in terms of aspects
and sentiments in reviews. 

\begin{figure*}[htb]
\scalebox{.92}{
 \begin{picture}(380,120)
 \put(42,4){\textsl{Las Vegas}}
  \put(160,4){\textsl{New York}}
  \put(290,4){\textsl{Los Angeles}}
    \put(420,4){\textsl{Miami}}
\put(0,70){\includegraphics[width=0.21\textwidth]{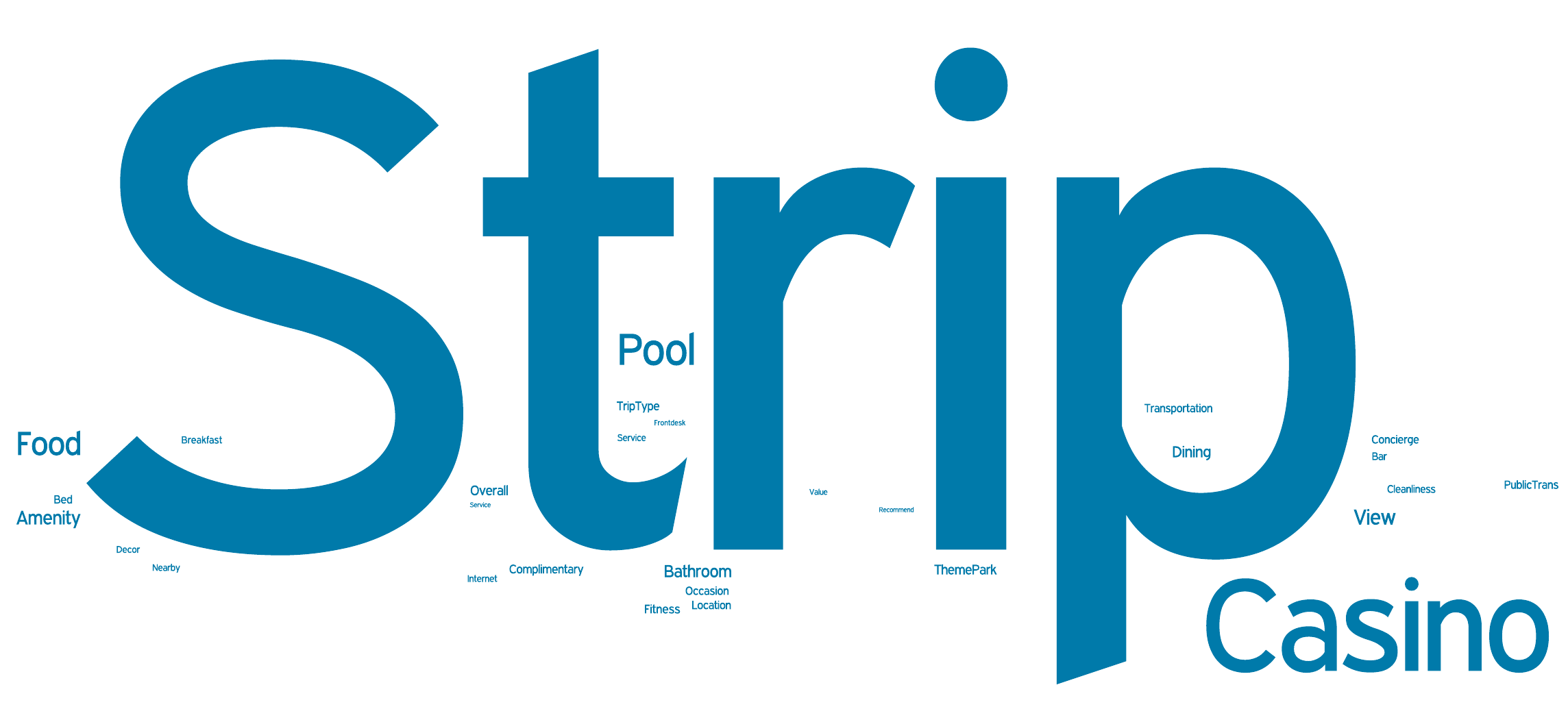}}
\put(120,70){\includegraphics[width=0.29\textwidth]{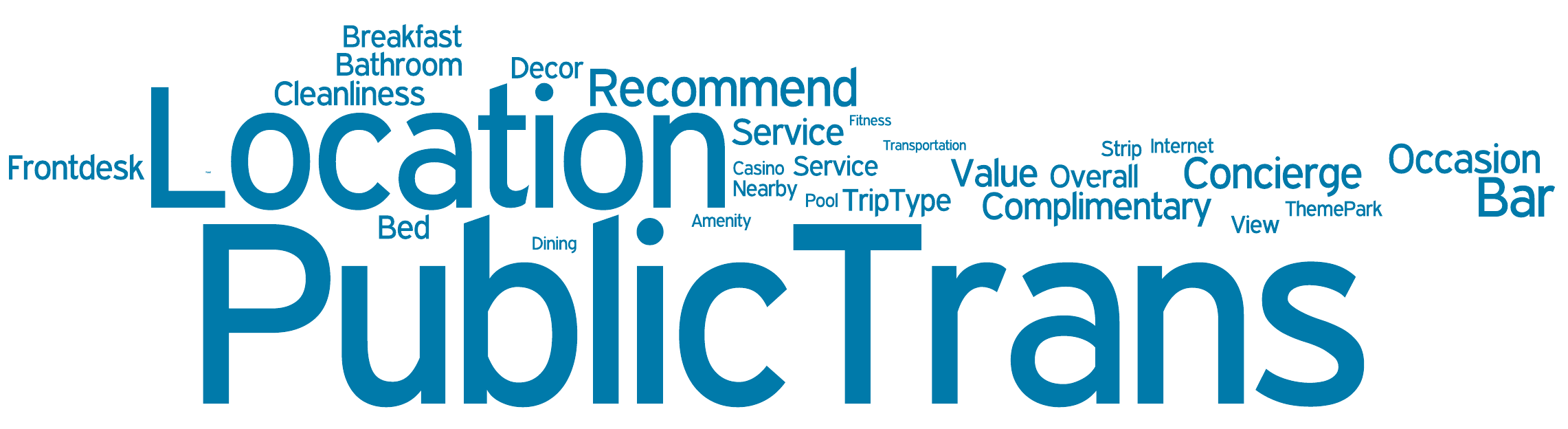}}
\put(250,78){\includegraphics[width=0.27\textwidth]{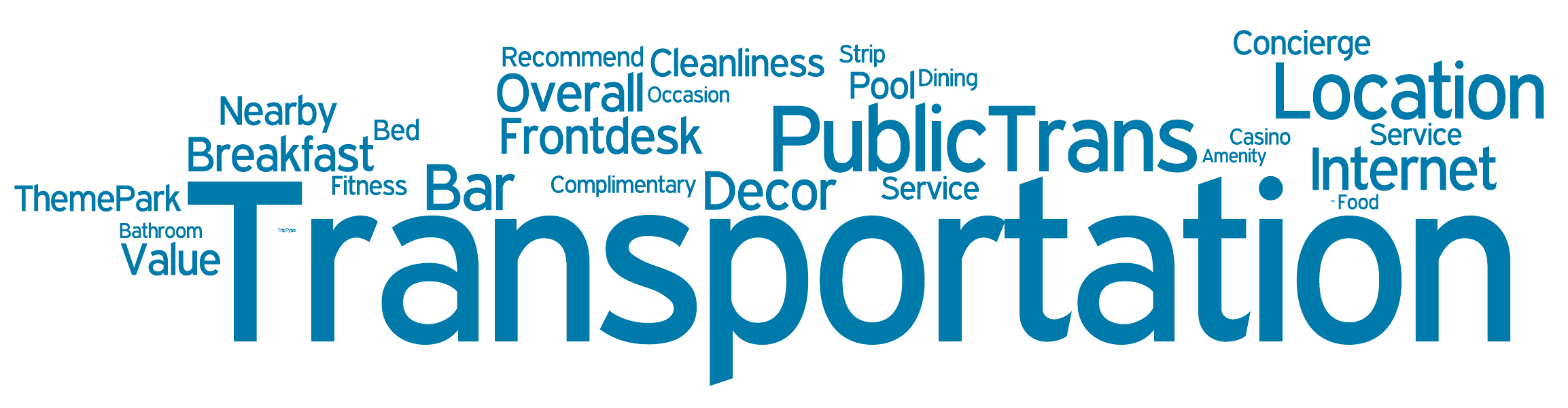}}
\put(378,83){\includegraphics[width=0.27\textwidth]{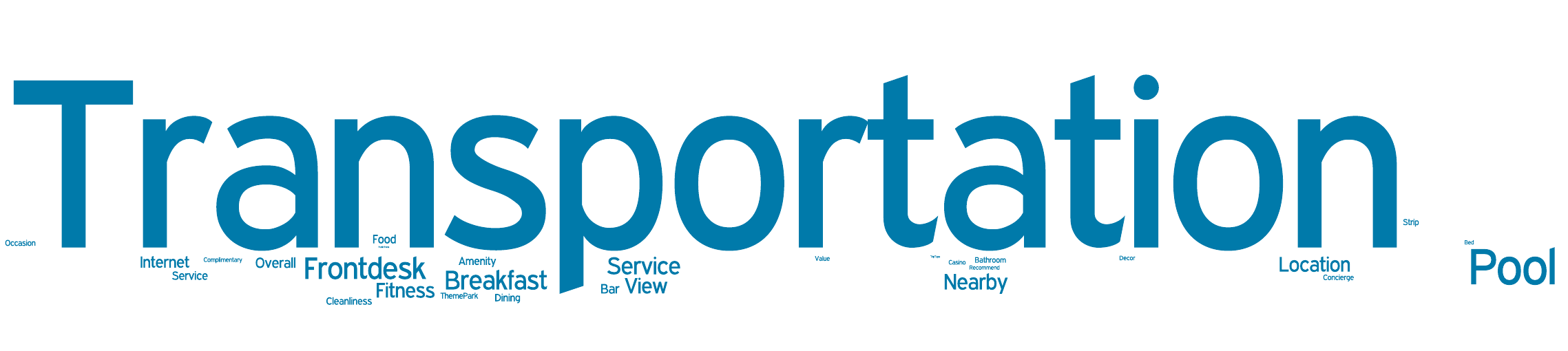}}

\put(0,20){\includegraphics[width=0.25\textwidth]{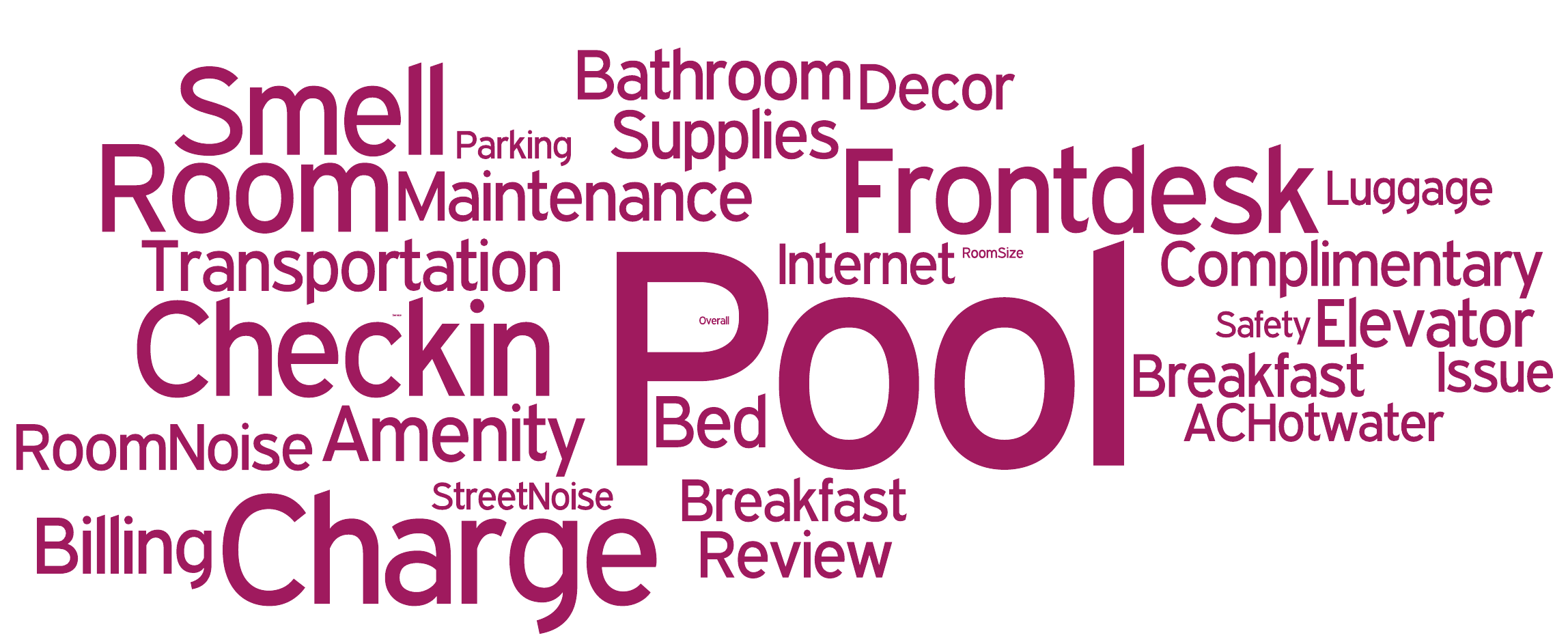}}
\put(120,20){\includegraphics[width=0.27\textwidth]{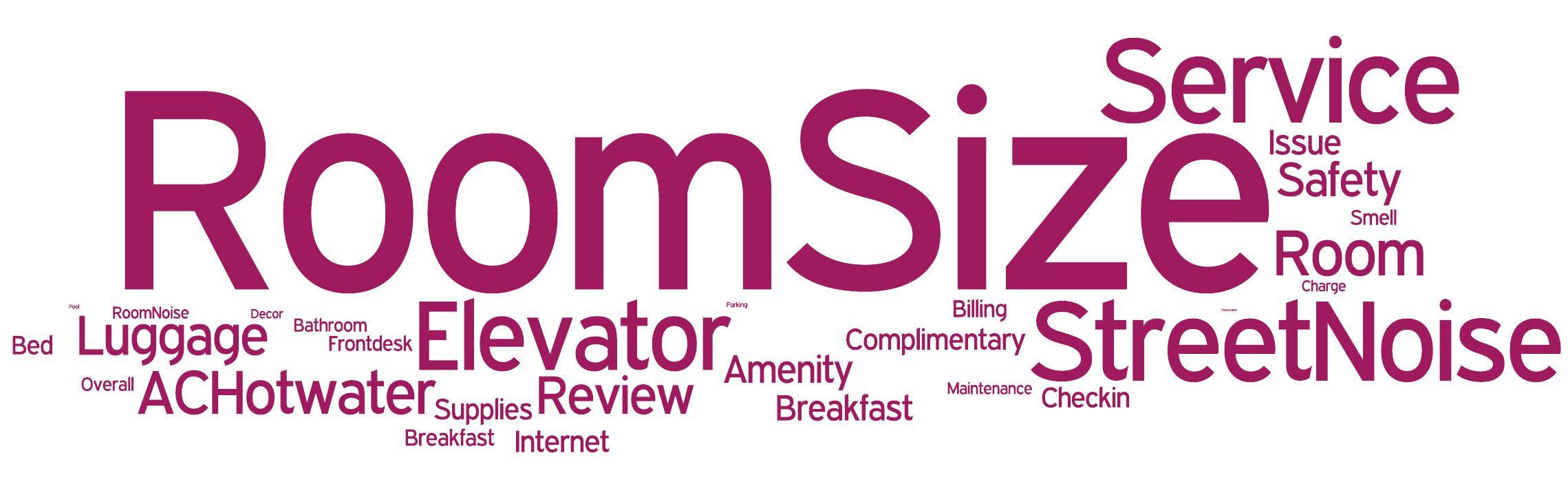}}
\put(250,15){\includegraphics[width=0.27\textwidth]{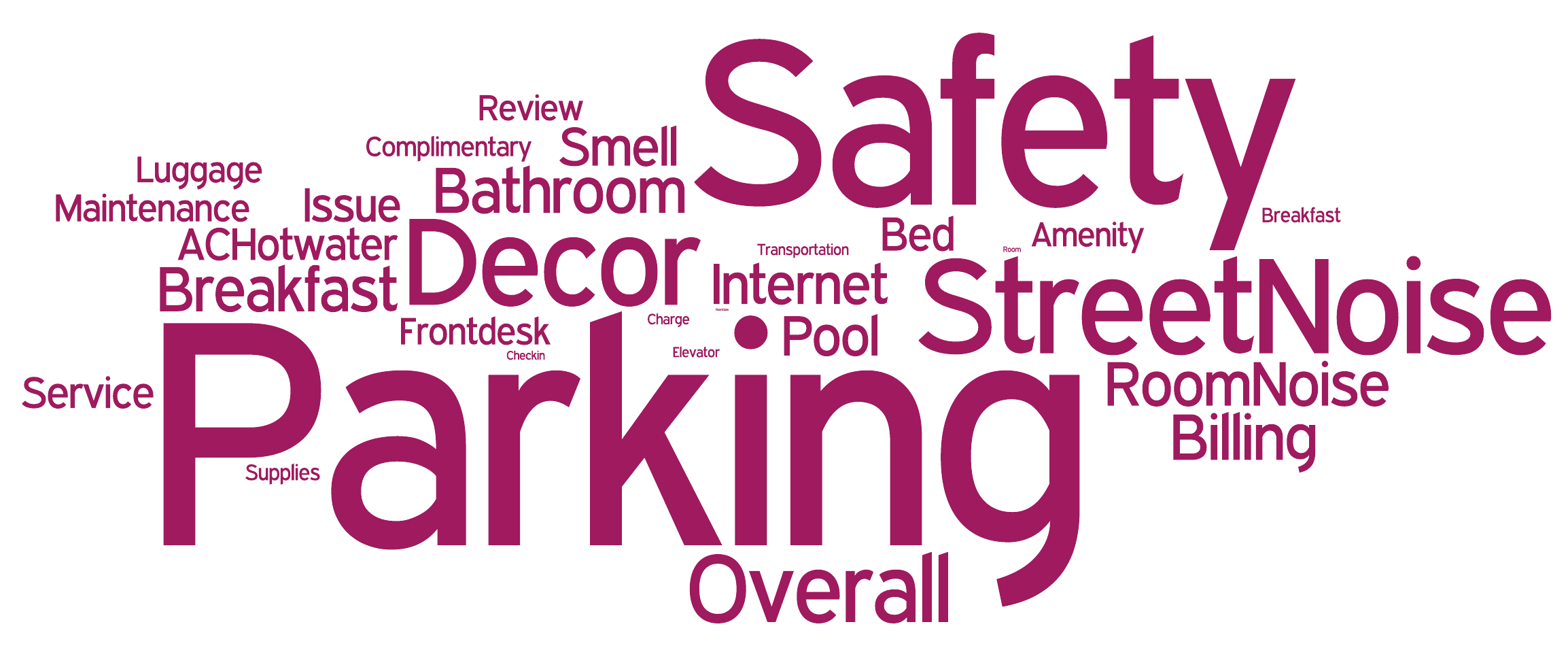}}
\put(378,20){\includegraphics[width=0.25\textwidth]{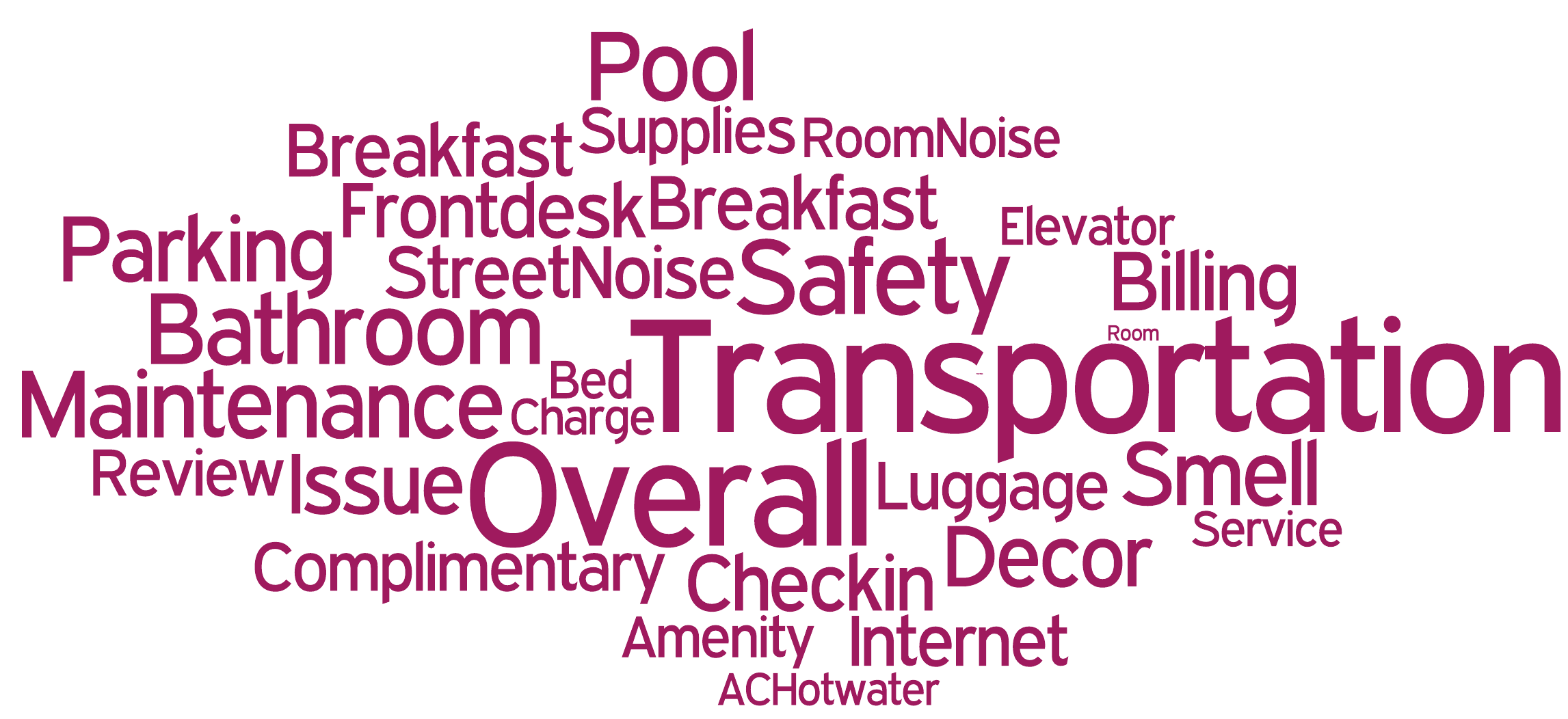}}
  \end{picture}
  }
\caption{An aspect-cloud visualization of US cities (positive aspects above; negative aspects below).}
\label{fig:profile_location}
\end{figure*}

\begin{figure*}[htb]
\scalebox{.92}{
 \begin{picture}(380,140)
 \put(42,4){\textsl{Business}}
  \put(160,4){\textsl{Couple}}
  \put(300,4){\textsl{Solo}}
    \put(420,4){\textsl{Family}}
\put(0,85){\includegraphics[width=0.28\textwidth]{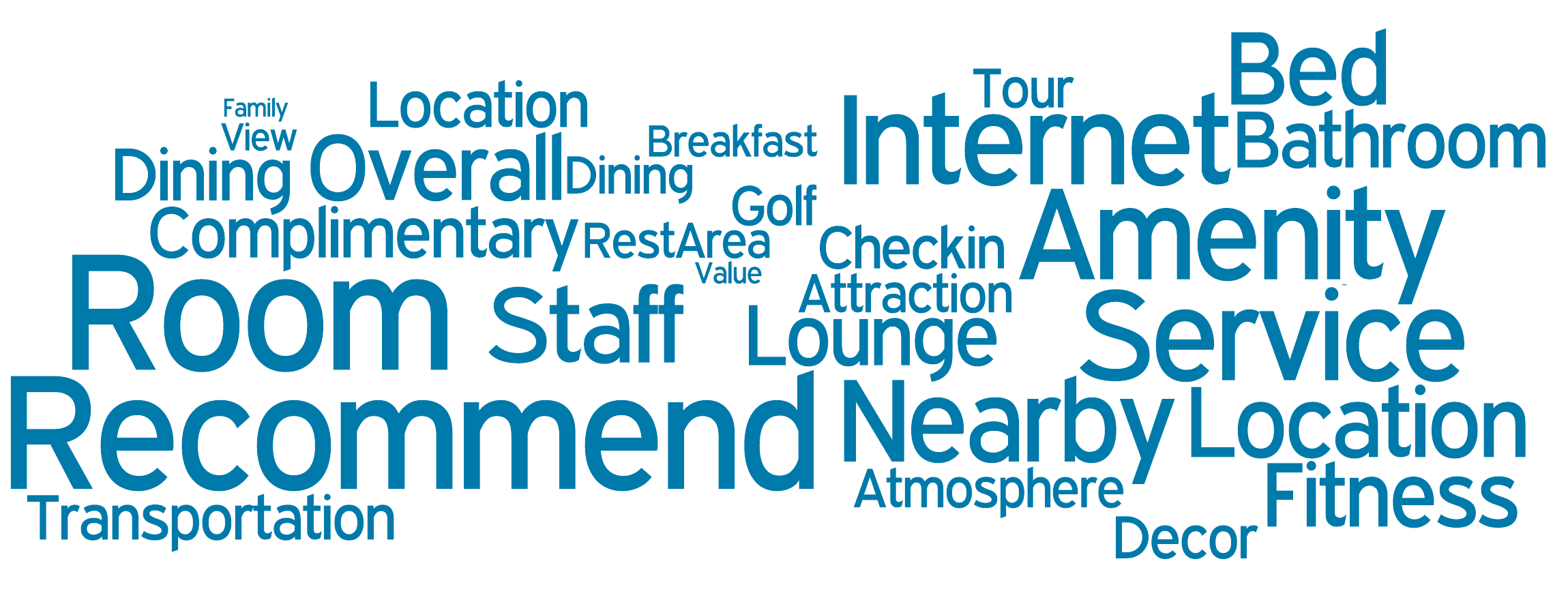}}
\put(127,85){\includegraphics[width=0.27\textwidth]{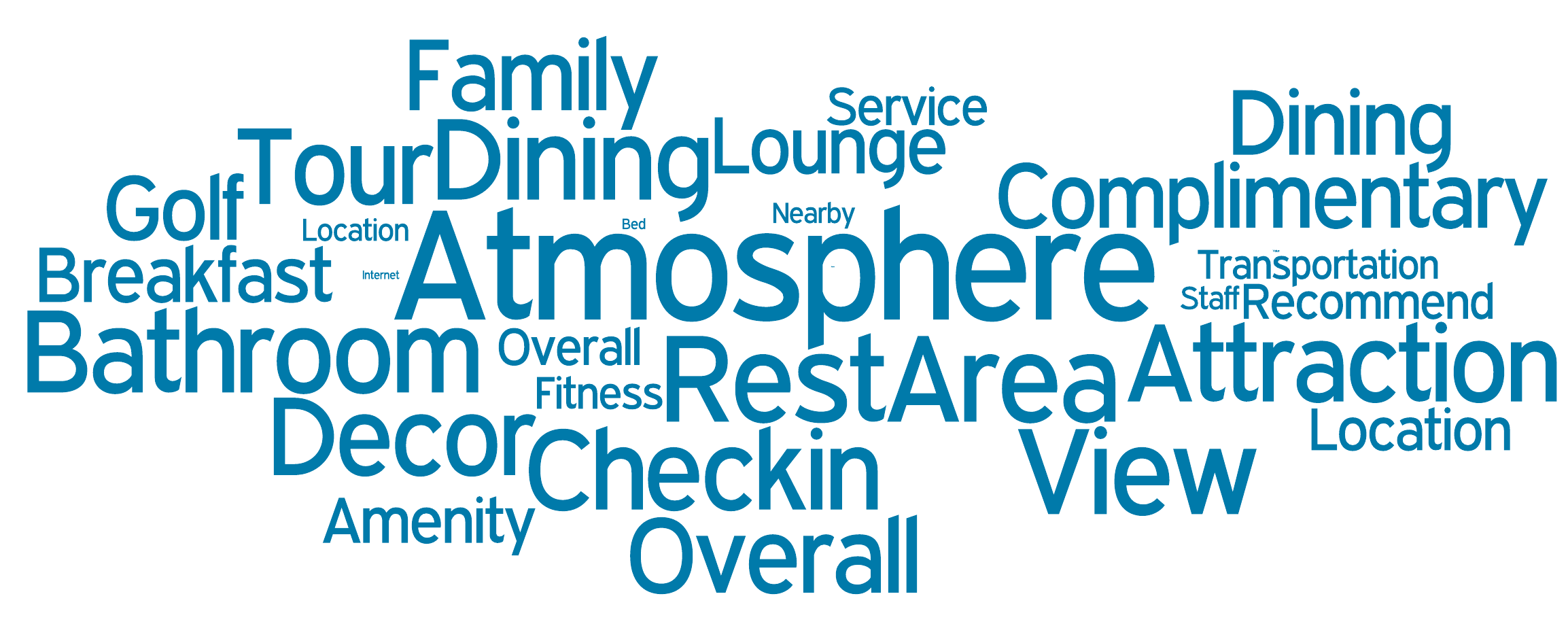}}
\put(250,85){\includegraphics[width=0.27\textwidth]{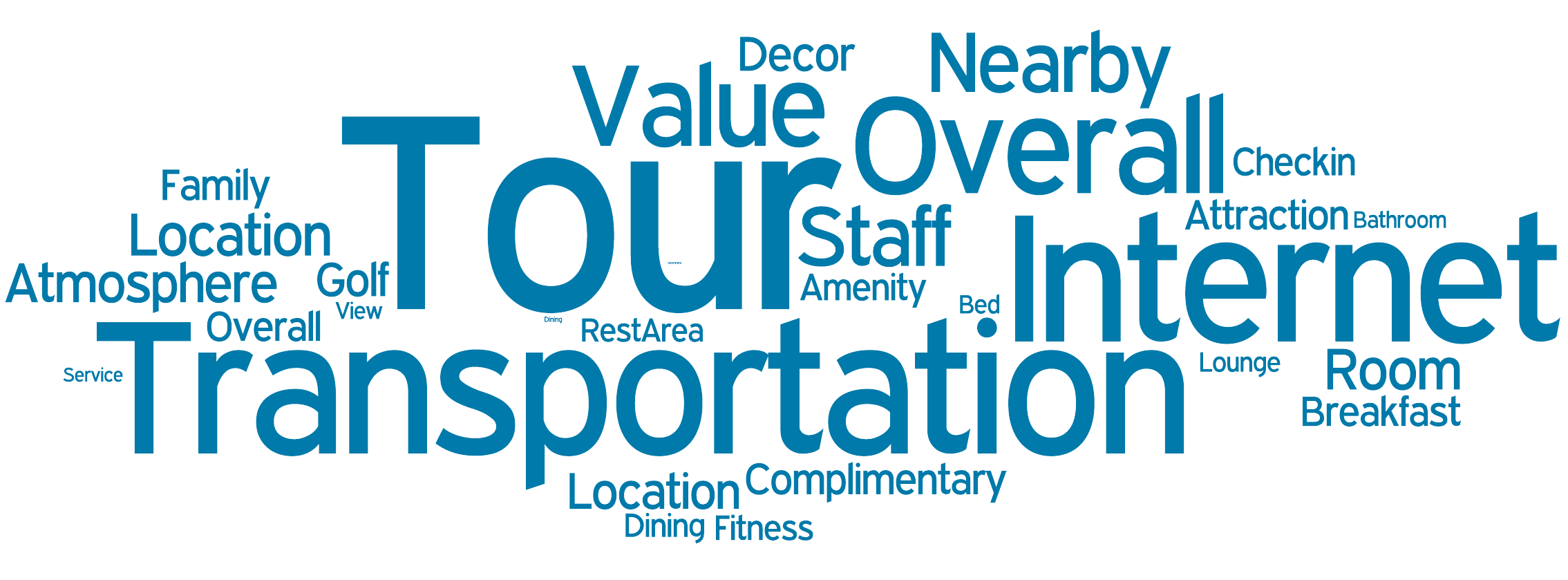}}
\put(378,85){\includegraphics[width=0.25\textwidth]{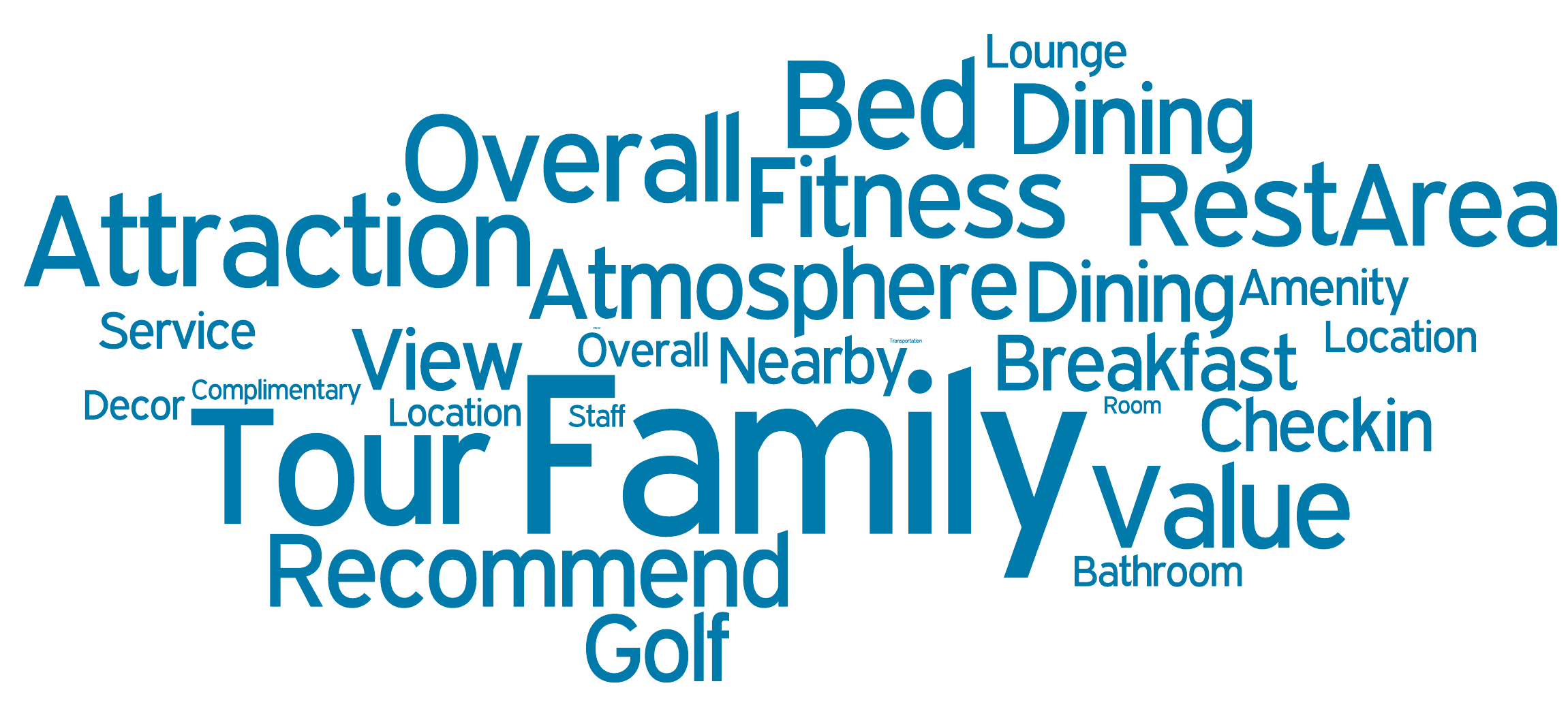}}

\put(-2,20){\includegraphics[width=0.28\textwidth]{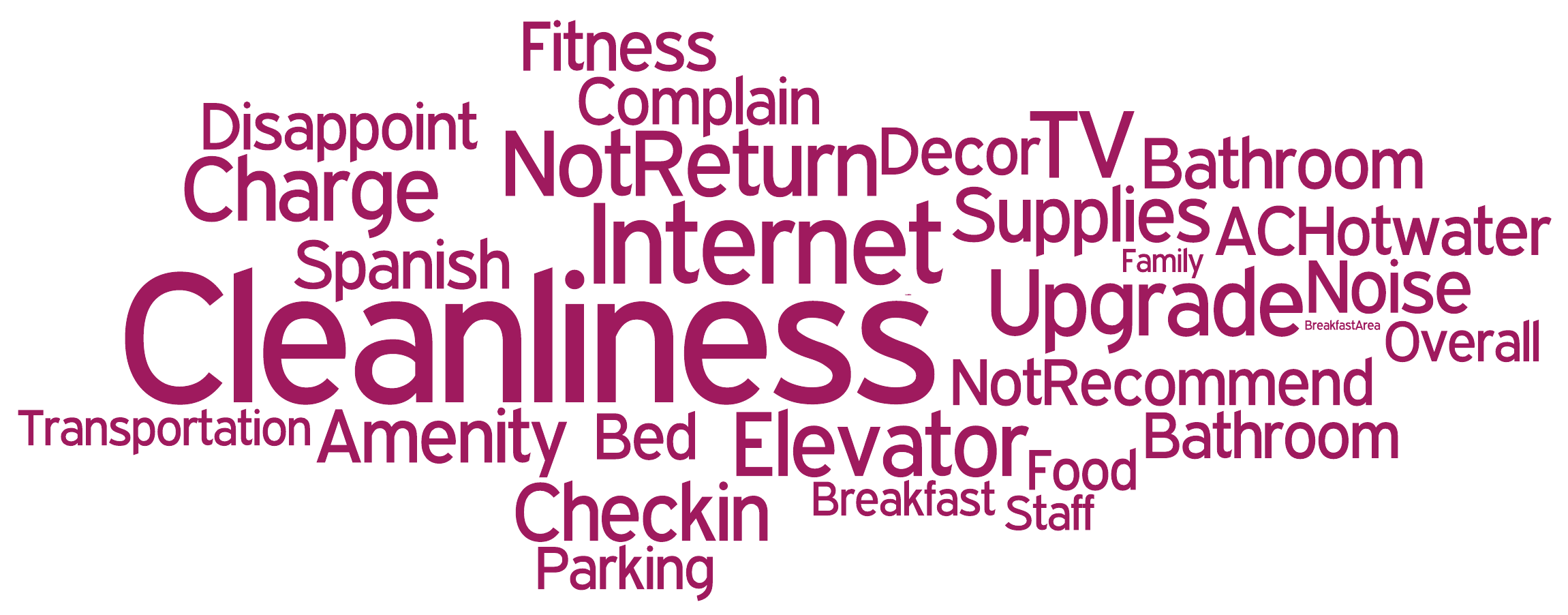}}
\put(120,20){\includegraphics[width=0.28\textwidth]{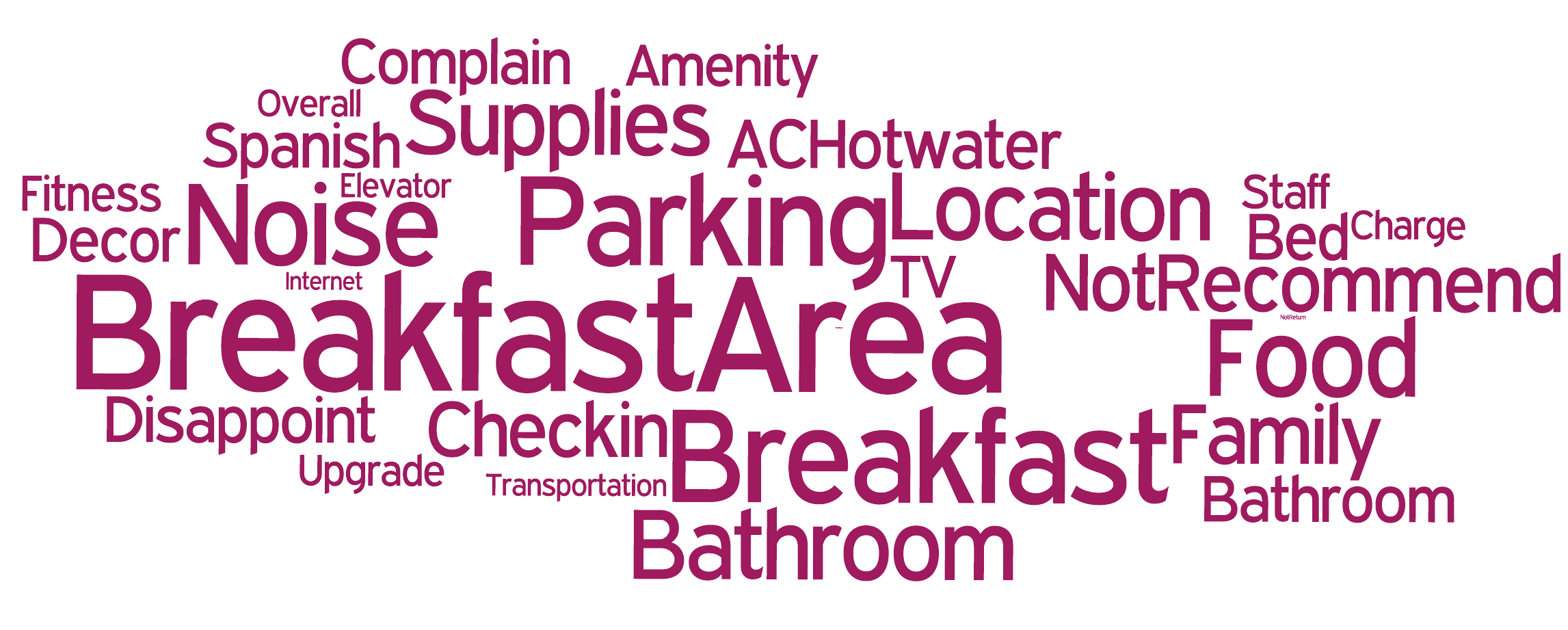}}
\put(250,20){\includegraphics[width=0.27\textwidth]{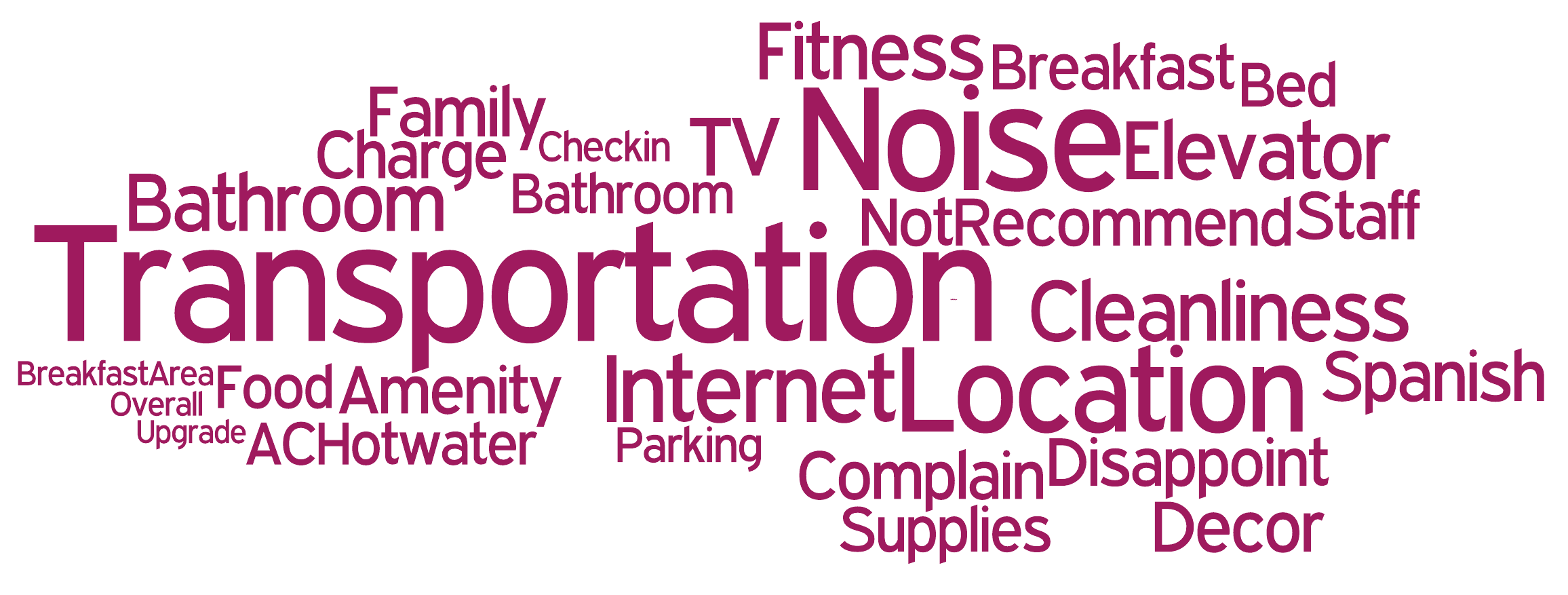}}
\put(378,20){\includegraphics[width=0.25\textwidth]{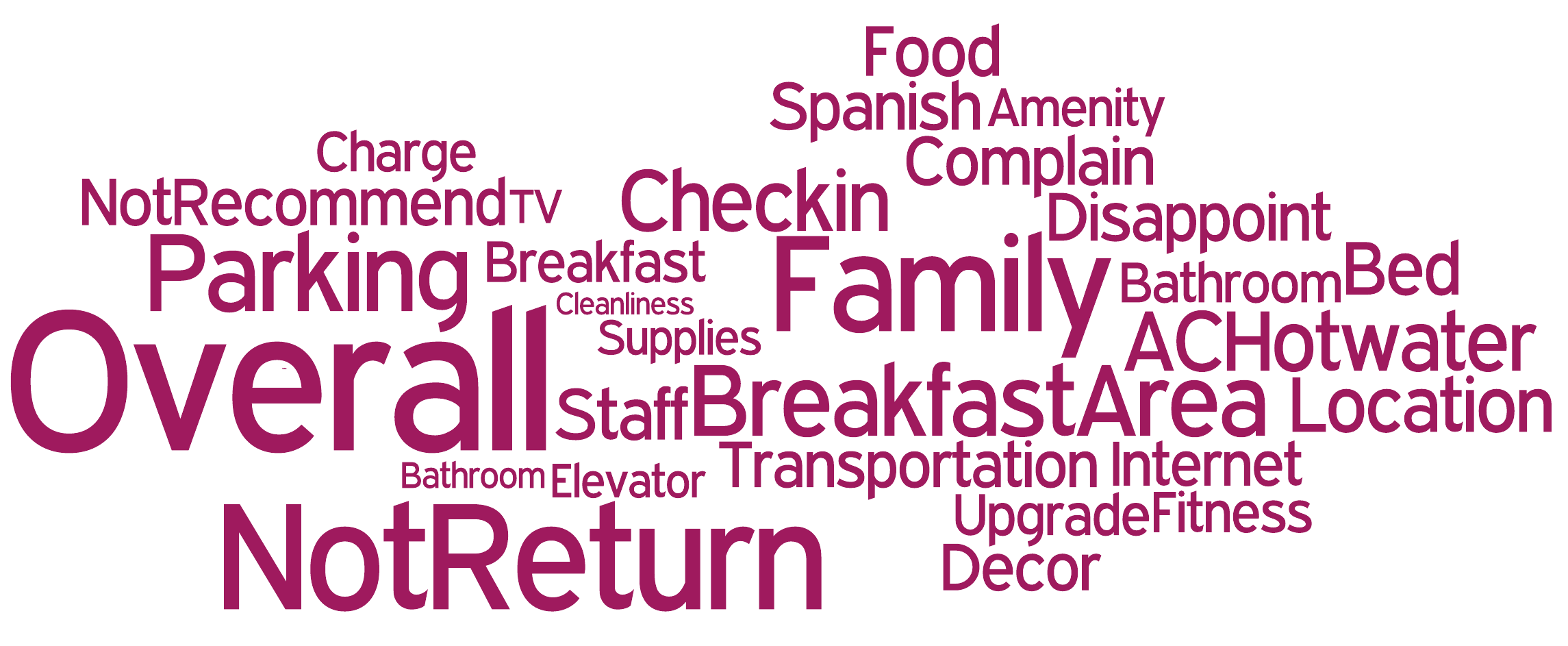}}
  \end{picture}
  }
\caption{An aspect-cloud visualization of trip types.  (positive aspects above; negative aspects below).}
\label{fig:profile_type}
\end{figure*}

Figure~\ref{fig:profile_location} shows the profiles of four US cities
generated by Trait. We visualize the profiles as aspect-clouds using
the top 30 aspects in each sentiment. The size of the aspect label
corresponds to its aspect probabilities. Due to space constraints, we
place the profiles of Boston, Chicago, and Orlando in
Table~\ref{tab:remain}.

\begin{table}[htb]
\centering
\caption{Top five aspects discovered by Trait.}
\label{tab:remain}
\scalebox{0.76} {
\begin{tabular}{@{~~} c@{~~~} c@{~~~} c@{~~~} c@{~~}}\toprule
\colu{Boston (P)}&\colu{Boston (N)}&\colu{Chicago (P)}&\colu{Chicago (N)}\\\midrule
PublicTrans&RoomSize&Location&RoomSize\\
Location&Service&PublicTrans&Service\\
Decor&Parking&Decor&Parking\\
Bar&RoomNoise&Bar&Elevator\\
Service&StreetNoise&Service&ACHotwater\\\midrule
\colu{Orlando (P)}&\colu{Orlando (N)}&\colu{Friend (P)}&\colu{Friend (N)}\\\midrule
ThemePark&Pool&Tour&Upgrade\\
Pool&Transportation&Value&Staff\\
Food&Supplies&View&BreakfastArea\\
Amenity&Frontdesk&Breakfast&Bathroom\\
Overall&Bed&Location&Breakfast\\\midrule
\colu{Author A (P)}&\colu{Author A(N)}&\colu{Author B (P)}&\colu{Author B (N)}\\\midrule
Helpfulness&NotReturn&Helpfulness&NotReturn\\
View&Value&View&Value\\
Value&Charge&Breakfast&Checkin\\
Breakfast&Checkout&Value&HotelSize\\
Comfort&Checkin&Transportation&Checkout\\\midrule
\colu{Author C (P)}&\colu{Author C (N)}&\colu{Author D (P)}&\colu{Author D (N)}\\\midrule
Internet&Elevator&Staff&Value\\
Comfort&Room&Room&NotReturn\\
TripType&Food&Recommend&Checkout\\
View&StreetNoise&Value&Room\\
Breakfast&Checkout&TripType&Charge\\
\bottomrule
\end{tabular}
}
\end{table}

These profiles yield salient summaries for each city. For example,
\emph{Strip} and \emph{Casino} are the top two positive aspects for
Las Vegas, a resort city for gambling. We see from the reviews that
most hotels with high ratings are located on the Strip. For Boston,
Chicago, and New York, \emph{Location} and \emph{PublicTrans} are the
top positive aspects. These cities rank top on the lists of U.S. cities with high transit ridership \cite{transit_ridership:2019} and walkability \cite{walk_rank:2019}. Hotels' proximity to public transportation, shopping, restaurants, and
attractions is appealing to several reviewers. We see \emph{RoomSize}
appears in the top five negative aspects. These three cities have
among the most expensive hotel room rates \cite{Statista:2019}.
Assuming consumers expect more when they pay more, we conjecture that
a failed expectation could be caused by room size, especially in New
York, where room sizes are smaller than elsewhere in the US
\cite{NYC:2019}. For Miami, \emph{Transportation} is attractive,
presumably because many cruises depart from Miami.

Figure~\ref{fig:profile_type} shows the results for HotelType:
\emph{Cleanliness}, \emph{Internet}, \emph{TV}, and \emph{Upgrade} are
most likely to lead to a negative sentiment for business travelers.
For couples, \emph{Atmosphere} and \emph{RestArea} are most preferred
positive aspects. \emph{Family}, \emph{Tour}, and \emph{Attraction}
are most positive aspects for families. Solo travelers, on business or
tourism, express most opinions toward both \emph{Transportation}.

Table~\ref{tab:remain} (bottom rows) lists the top five aspects for
four authors from HotelUser. We can observe strong commonality between
Author A and B, They both like to express positive sentiment on
\emph{Helpfulness}, \emph{View}, \emph{Value}, and \emph{Breakfast}.
They are like to express not returning a hotel and the negative
sentiments are mostly toward \emph{Value}, \emph{Checkin}, and
\emph{Checkout}. There is little commonality between Authors C and D.
For Author C, \emph{Internet} and \emph{Comfort} are most attractive
whereas \emph{Staff} and \emph{Room} are most appealing aspects for
Author D. In terms of negatives, \emph{Checkout} is the only aspect shared
between Authors C and D.

\subsubsection{Similarity}

Attribute profiles can be used not only for summarization, but also
for measuring similarity between attribute values with respect to
aspects and sentiments, which can support attribute-based applications
such as recommender systems. Our metric of similarity between distinct
values of the same attribute is the Jensen-Shannon distance (JSD)
\cite{EndresS:03}, the square root of Jensen-Shannon divergence. We
compute, $D_{\text{JS}}$, the JSD of attribute profiles $P$ and $Q$ as

{\setlength{\mathindent}{0cm}
\begin{flalign}
\label{eq:JSD}
\begin{split}
D_{\text{JS}} &= \sqrt{\frac{1}{2} D_{KL} (P||M) + \frac{1}{2} D_{KL}(Q||M)} \\
M &= \frac{1}{2}(P+Q),
\end{split}
\end{flalign}
}

{\noindent}
where $D_{KL} (P||Q)$ is the Kullback-Leibler (KL) divergence of probability distributions $P = \{p_1,\dots,p_n\}$ and $Q=\{q_1,\dots,q_n\}$:

\begin{align}
\label{eq:KL}
\begin{split}
D_{KL} (P||Q) = \sum_{i} p_i \log\frac{p_i}{q_i}.
\end{split}
\end{align}

As a baseline, we use a vector space model based on USE sentence
embeddings. We calculate mean sentence embeddings for each review.
Then, given two sets of reviews, $D=\{d_1,\dots,d_m\}$ and
$R=\{r_1,\dots,r_n\}$, we compute their similarity as follows (here
$sim(d_i, r_j)$ is the cosine similarity between review $d_i$ and
$r_j$):

\begin{align}
S_{D,R} =\frac{1}{m+n}  \sum_{i=1}^{m}  \sum_{j=1}^{n} sim(d_i, r_j).
\end{align}

\begin{figure}[htb]
\centering
\begin{subfigure}[t]{0.82\linewidth}
\includegraphics[width=1\textwidth]{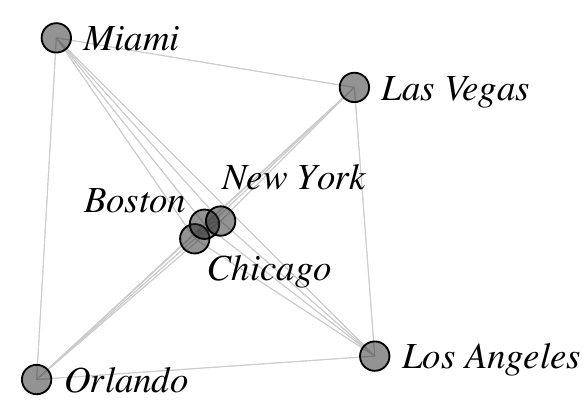}\caption{Baseline.}
\label{fig:city_dist_base}
\end{subfigure}
\hfill
\begin{subfigure}[t]{0.85\linewidth}
\includegraphics[width = 1\textwidth]{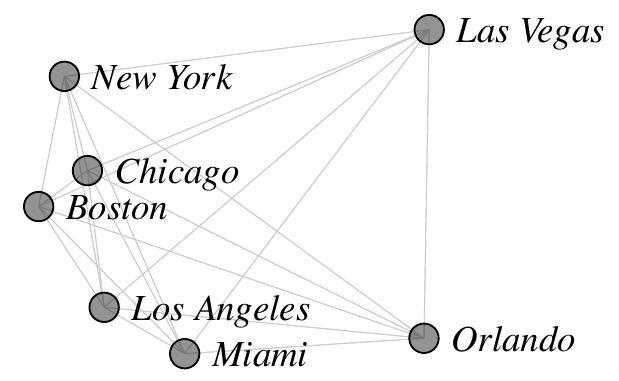}
\caption{Trait.}
\label{fig:city_dist_Trait}
\end{subfigure}
\caption{Force fields of similarities: cities.}
\label{fig:city_dist}
\end{figure}

Figure~\ref{fig:city_dist_base} shows the similarities among the
profiles of the seven cities generated by the baseline model. We see
that Boston is close to New York and Chicago; Las Vegas is far away
from Los Angeles and Miami but close to New York, Boston, and Chicago.
Figure~\ref{fig:city_dist_Trait} shows the results generated by Trait.
Here, dissimilarity corresponds to distance normalized to $[0, 1]$.
Boston, Chicago, and New York are close to each other, as are Los
Angeles and Miami; Las Vegas is far from each of the others; and,
Orlando is far from all except Miami. Trait's results are arguably
more plausible than what the baseline approach produces.

As discussed earlier, hotels in Boston, Chicago, and New York have
common characteristics; Las Vegas differs strongly from the others
because it is a resort city and its major hotels are combined with
casinos; Orlando is a tourism destination but differs from Las Vegas
in that it is famous for local attractions, such as theme parks.
Orlando's profile exhibits that travelers there tend to be more aware
of aspect \emph{Attraction} than elsewhere. An interesting pair is Los
Angeles and Miami. We see that \emph{Location} appears as the most
important aspect on the positive side for both of them. Such a
similarity could be partially explained by the fact that both Los
Angeles and Miami serve as locations for taking cruises. Also, the
common aspect \emph{Safety} (negative for both) could increase their
similarity.

\begin{figure}[htb]
\centering
\begin{subfigure}[t]{0.75\linewidth}
\includegraphics[width=1\textwidth]{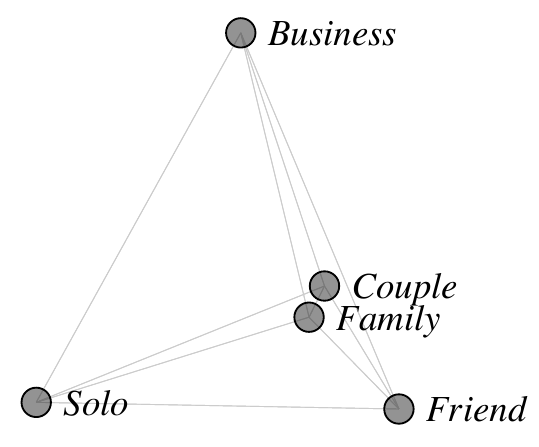}
\caption{Baseline.}
\label{fig:type_dist_base}
\end{subfigure}
\hfill
\begin{subfigure}[t]{0.88\linewidth}
\includegraphics[width = 1\textwidth]{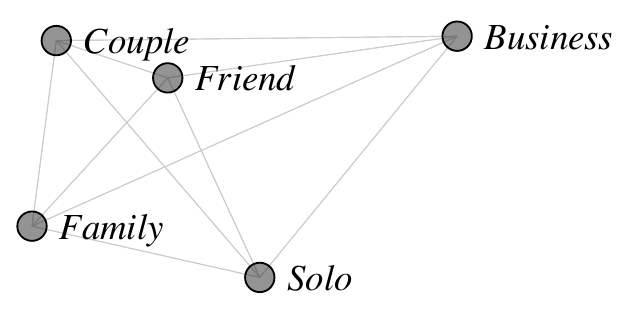}
\caption{Trait.}
\label{fig:type_dist_Trait}
\end{subfigure}
\caption{Force fields of similarities: Purpose.}
\end{figure}

Figure~\ref{fig:type_dist_base} shows similarities among the five trip
types generated by the baseline model. Business is closer to Family
than Solo and Friend; Solo is closer to Family than Friend. Trait
generates more reasonable results, as shown in
Figure~\ref{fig:type_dist_Trait}. Business is far from others but is
closer to Friend and Solo than to Couple and Family. Couple, Family,
and Friend are relatively close to each other. Business reviewers
attend to different aspects from other reviewers. Further, Solo and
Friend contain reviews of business trips, although the authors did not
select Business as the trip type. However, this situation does not
happen for Couple and Family.

\begin{figure}[htb]
\centering
\includegraphics[width=0.42\textwidth]{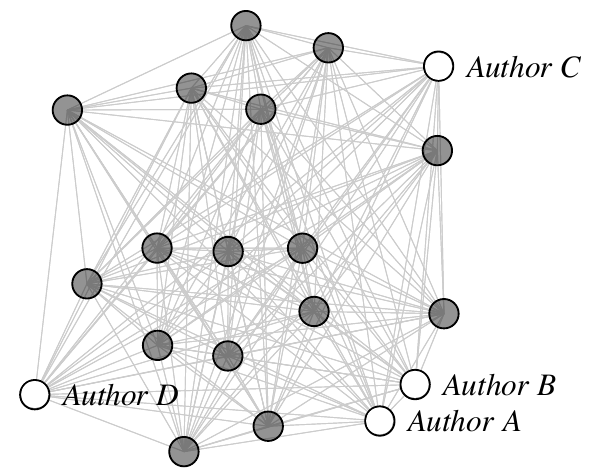}
\caption{Force field of similarities: 20 authors. }
\label{fig:user_dist}
\end{figure}

Figure~\ref{fig:user_dist} shows similarities among 20 authors,
including the four authors mentioned in
Section~\ref{subsubsec:summarization}. We see that Authors A and B are
close to each other whereas Authors C and D are far away from each
other. The results are aligned with their aspect and sentiment
profiles.

\section{Discussion and Conclusion}

Trait not only shows that capturing structural and semantic
correspondence leads to improved performance in terms of coherence and
naturalness of the aspects discovered but can also be realized in an
unsupervised framework. Trait outperforms competing approaches across
multiple datasets.

These results open up interesting directions for future work. One
direction is to learn disentangled latent representations for attributes in neural network's space, such as for disentangling aspects \cite{jain:2018}, text style \cite{john:2019}, and syntax and semantics \cite{chen:2019, bao:2019}. Another direction is to develop a content-based recommender based on Trait,
since it provides an effective unsupervised solution for generating
profiles based on different attributes. 


\section{Acknowledgments}
We would like to thank the anonymous reviewers for helpful comments and corrections.
MPS would like to acknowledge the US Department of Defense for partial support through the NCSU Laboratory for Analytic Sciences.

\bibliographystyle{acl_natbib}

\end{document}